\documentclass{article}

\usepackage[preprint]{neurips_2022}

\usepackage[utf8]{inputenc} 
\usepackage[T1]{fontenc}    
\usepackage{hyperref}       
\usepackage{url}            
\usepackage{booktabs}       
\usepackage{amsfonts}       
\usepackage{nicefrac}       
\usepackage{microtype}      
\usepackage{xcolor}         

\usepackage{soul}
\usepackage{graphicx}
\usepackage{subfig}
\usepackage{enumitem}
\usepackage{amsmath}
\usepackage{bm}
\usepackage{bbm}
\usepackage{multirow}

\usepackage[english]{babel}
\usepackage{lmodern}
\usepackage{tabularx}

\title{The Fixed Sub-Center: A Better Way to Capture Data Complexity}

\author{%
  Zhemin Zhang, Xun Gong, Jinyi Wu \\
  Southwest Jiaotong University, China\\
  \texttt{zheminzhang@my.swjtu.edu.cn} \\
}

\begin{document}

\maketitle

\sethlcolor{yellow}
\setstcolor{red}
\setcitestyle{numbers}

\begin{abstract}
Treating class with a single center may hardly capture data distribution complexities. Using multiple sub-centers is an alternative way to address this problem. However, highly overlapping sub-classes, the classifier's parameters grow linearly with the number of classes, and lack of intra-class compactness are three typical issues that need to be addressed in existing multi-subclass methods. To this end, we propose to use Fixed Sub-Center (F-SC), which allows the model to create more discrepant sub-centers while saving memory and cutting computational costs considerably. The F-SC specifically, first generates a normal distribution for each class, and then samples the sub-centers based on the normal distribution corresponding to each class, and the sub-centers are fixed during the training process avoiding the overhead of gradient calculation. Finally, F-SC penalizes the Euclidean distance between the samples and their corresponding sub-centers, it helps remain intra-compactness. The experimental results show that F-SC significantly improves the accuracy of both image classification and fine-grained recognition tasks.
\end{abstract}

\section{Introduction}

Image classification has made significant progress in recent years \cite{he2016deep,liu2022convnet}. Image classification includes several classification tasks, such as face recognition, target detection, and fine-grained image classification. With the development of deep learning, deep convolutional neural networks (DCNNs) have gradually become the most preferred method for image classification. Based on large-scale image data, the image feature embeddings (hereinafter referred to as features) are extracted by neural networks, which are free from the traditional manual design method of extracting features. Research shows that neural networks can better use large-scale image data to extract more discriminative features.

\begin{figure}[tb]
   \centering
   \includegraphics[width=1.0\linewidth]{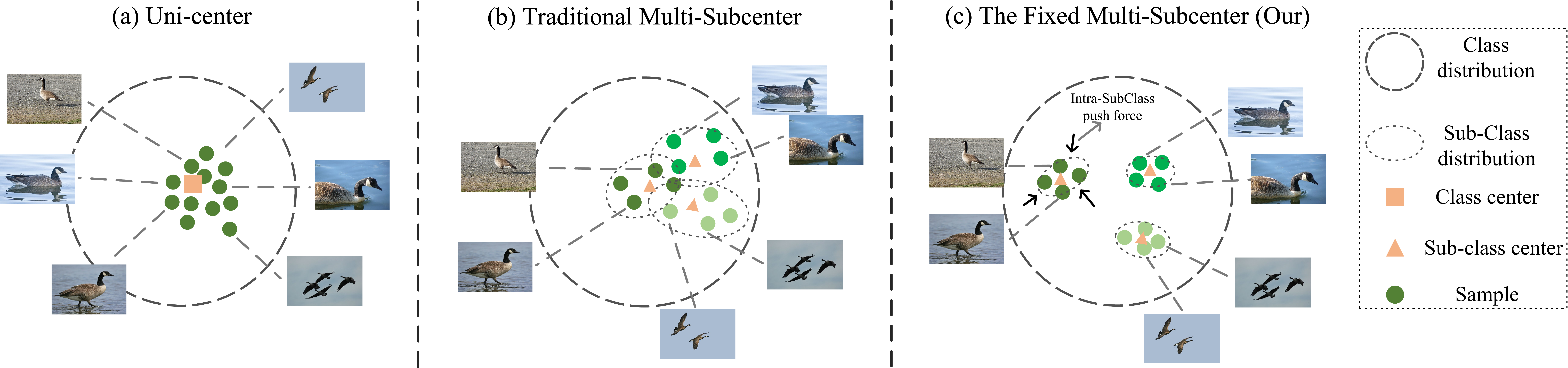}\\
   \caption{Comparison of uni-center, traditional multi-subcenter, and our fixed multi-subcenter when the class has different sub-classes. (a) Samples in the same class will be gathered to the same center. It may be inappropriate for the real-world data. (b) In contrast, it is more flexible for modeling intra-class variance by setting multiple sub-centers in a class, but traditional multi-subcenter methods may have the problems of lack of diversity of sub-centers and reduced intra-subclass compactness. (c) The Fixed Sub-Center improves the intra-subclass compactness while ensuring the diversity of sub-centers.}
   \label{fig1}
\end{figure}

DCNNs consist of a stack of convolutional parameterized layers, spatial pooling layers and fully-connected layers. The last fully-connected layer (classiﬁer) is usually regarded as a classifier, transforming from the dimension of network features $d$ to the number of required class categories $C$. The classiﬁer's outputs are class score vectors, which are then input into the SoftMax for training, allowing the model extract features with a high discriminative ability \cite{deng2019arcface,pmlr-v139-yang21o}. Research shows \cite{qian2019softtriple,9684999} that the classiﬁer provides a classification center for each class. The distance between each sample's feature and its corresponding class center is optimized during the training process using SoftMax. However, as shown in Figure \ref{fig1}, classes in real-world data generally do not cluster in a single center, but are prone to multiple local sub-centers. Therefore, a single center may be not representable enough for data distribution. The feature extracted by networks trained with a single center loses some diversity compared to the raw data distributions \cite{sohn2016improved,GUO2021104033}.

Recently, researchers suggested that a class can be divided into multiple sub-classes by setting multiple sub-centers \cite{qian2019softtriple,DBLP:journals/corr/abs-2112-11689,10.1145/3474085.3475536}. However, three issues need to be addressed to utilize multiple sub-centers fully: \textbf{(1)} Highly overlapping sub-centers. The sub-centers are close to each other during training and eventually overlap. Highly overlapping sub-centers cannot capture the sub-classes with different modalities. \textbf{(2)} The classifier has a vast number of trainable parameters. For the multiple sub-centers methods where the number of classes is $C$, the feature dimension is $d$, and each class has $S$ sub-classes, the classification model must hold $C\cdot d\cdot S$ number of trainable parameters that grows linearly with the number of classes and the number of sub-classes. \textbf{(3)} Undermine the intra-class compactness. Dividing a class into multiple sub-classes may undermine the intra-class compactness of features, which is important for classification.

We propose to use Fixed Sub-Center (F-SC), generating sub-classes with more diversity and reduce the amount of calculation, to address these problems above. Meanwhile, the Euclidean distance between samples and their corresponding sub-centers is penalized to ensure intra-subclass compactness. Our method is easy to be applied to other off-the-shelf models by making a few small changes in the classiﬁer.

\begin{figure}[tb]
   \centering
   \includegraphics[width=1.0\linewidth]{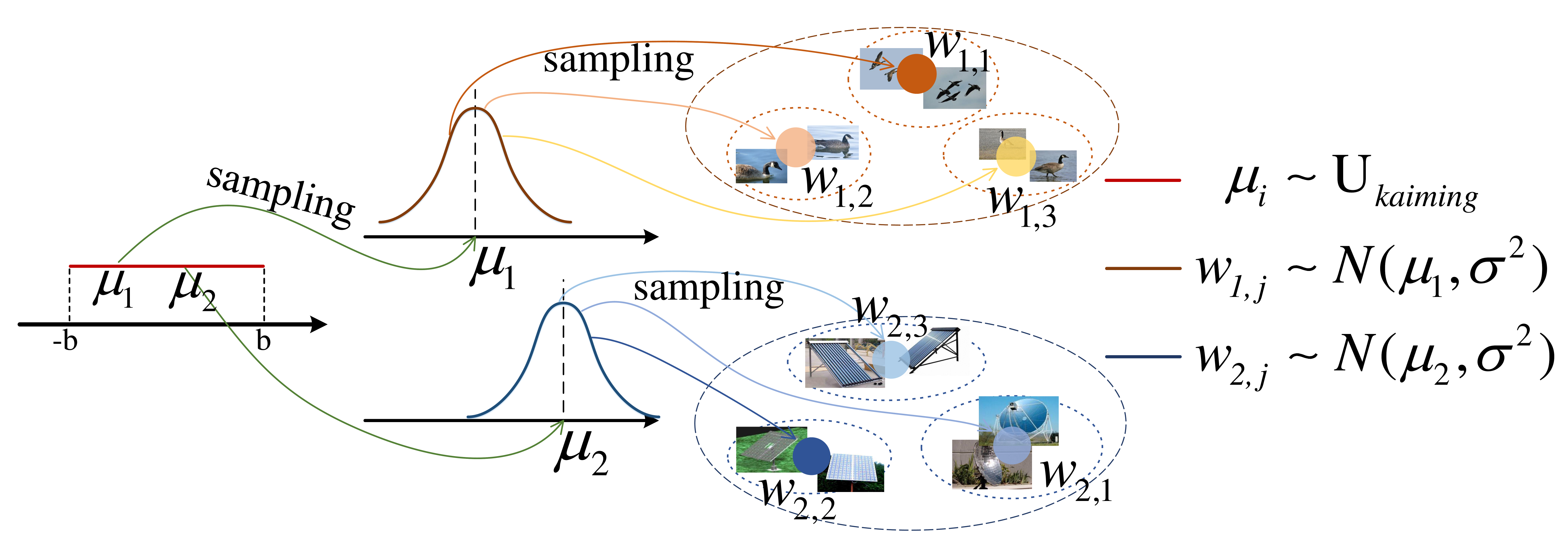}\\
   \caption{Basic idea of the proposed F-SC. Follow the classifier initialization method in kaiming-uniform \cite{He_2015_ICCV} to sample a class center ${{\mu }_{i}}$ for each class, and then generate a normal distribution for each class, where the mean is equal to ${{\mu }_{i}}$. Finally, the sub-centers are sampled based on the normal distribution for each class.}
   \label{fig2}
\end{figure}

In summary, our main contributions are as follows:
\begin{enumerate}
   \item Sample the sub-centers based on a normal distribution, as shown in Figure \ref{fig2}. The distance between the sub-centers can be adjusted by setting different variances, generating more diverse sub-classes.
   \item The parameters of the classiﬁer are fixed during the training process, keeping the initial sampling values unchanged and no gradient computation is required. This can greatly reduce the amount of extra calculations in the traditional multiple sub-centers methods.
   \item The F-SC penalizes the Euclidean distance between the samples and their corresponding sub-centers, which helps recapture intra-compactness.
\end{enumerate}

\section{Method}

SoftMax, optimizing by gradient descent, reduces the distance between each sample’s feature and the weights of its corresponding classiﬁer. The weight of each class in the classiﬁer can be regarded as the class center. The proposed F-SC aims to capture the sample distribution by setting multiple sub-centers for each class. 

In this part, we first introduce SoftMax loss and then define our F-SC loss. The feature of the $i$-th sample is denoted as ${{\boldsymbol{x}}_{i}}$, and the corresponding label is denoted as ${{y}_{i}}$. DCNNs calculate the probability that ${{\boldsymbol{x}}_{i}}$ belongs to ${{y}_{i}}$ by SoftMax as:
\begin{equation} 
{{P}_{{{\boldsymbol{x}}_{i}},{{y}_{i}}}}=\frac{\exp (\boldsymbol{w}_{{{y}_{i}}}^{\rm T}{{\boldsymbol{x}}_{i}})}{\sum\nolimits_{j=1}^{c}{\exp (\boldsymbol{w}_{j}^{\rm T}{{\boldsymbol{x}}_{i}})}} \label{1} 
\end{equation}
where $[{{\boldsymbol{w}}_{1}},\cdots ,{{\boldsymbol{w}}_{c}}]\in {{R}^{d\times c}}$ is the weights in the classiﬁer, $c$ denotes the number of classes, and $d$ denotes the dimensionality of the feature.

\subsection{Sub-Centers}
We set $s$ sub-centers for each class. Feature $\boldsymbol{x}\in {{R}^{d}}$ and all sub-centers $\boldsymbol{w}\in {{R}^{c\times s\times d}}$. We obtain the subclass-wise similarity scores $\ logi{{t}_{\boldsymbol{x}}}\in {{R}^{c\times s}}$ by matrix multiplication $\boldsymbol{x}\cdot \boldsymbol{w}$. With multiple sub-centers, ${{P}_{{{\boldsymbol{x}}_{i}},{{y}_{i}},k}}$ denotes the probability that sample ${{\boldsymbol{x}}_{i}}$ belongs to the $k$-th sub-class of class ${{y}_{i}}$. The output probability ${{P}_{{{\boldsymbol{x}}_{i}},{{y}_{i}},k}}$ is calculated in a similar way to Eq.(\ref{1}).
\begin{equation} 
{{P}_{{{\boldsymbol{x}}_{i}},{{y}_{i}},k}}=\frac{\exp (\boldsymbol{w}_{{{y}_{i}},k}^{\rm T}{{\boldsymbol{x}}_{i}})}{\sum\nolimits_{j=1}^{c}{\sum\nolimits_{m=1}^{s}{\exp (\boldsymbol{w}_{j,m}^{\rm T}{{\boldsymbol{x}}_{i}})}}} \label{2} 
\end{equation}
where ${{\boldsymbol{w}}_{j,m}}$ denotes the $m$-th sub-center of class $j$. The sum of all sub-class probabilities of a class equals the prediction probability of the class. The models are trained by minimizing cross-entropy:
\begin{equation} 
{{L}_{sub-center}}=-\frac{1}{n}\sum\limits_{i=1}^{n}{\log (\sum\limits_{k=1}^{s}{{{P}_{{{\boldsymbol{x}}_{i}},{{y}_{i}},k}}})} 
\label{3} 
\end{equation}
where $n$ denotes the number of samples in a mini-batch.

\subsection{Using A Fixed Parameter Classifier}

As shown in Figure \ref{fig1}(b), after setting multiple sub-centers for each class, we need to ensure that the sub-centers do not overlap each other during training. If there is no difference between sub-centers of the same class, the goal of using multiple sub-centers cannot be fulfilled. Meanwhile, we hope that setting multiple sub-centers will not lead to a linear increase in the amount of calculation during model training. FYC \cite{hoffer2018fix} proves that for common use-cases of convolutional network, the parameters used for the ﬁnal classiﬁcation transform are completely redundant, and can be replaced with a pre-determined linear transform.

Kaiming-uniform \cite{He_2015_ICCV} is the default initialization method of ‘nn.Linear’ in PyTorch. For a fair comparison with other multi-subcenter methods, F-SC uses kaiming-uniform to sample the mean of the normal distribution:

\begin{equation} 
\mu =[{{\mu }_{1}},{{\mu }_{2}},{{\mu }_{3}},\cdots ,{{\mu }_{c}}],{{\mu }_{i}}\sim {{\mathsf{U}}_{\mathrm{kaiming}}}
\label{4} 
\end{equation}
where ${{\mu }_{i}}\in {{R}^{d}}$ denotes the center of the $i$-th class. We assume that the distribution of features of each class follows a normal distribution, and generate a corresponding normal distribution $\mathsf{N}({{\mu }_{i}},{{\sigma }^{2}})$ for each class through the class centers initialized in Eq.(\ref{4}). Through $\mathsf{N}({{\mu }_{i}},{{\sigma }^{2}})$, we can sample multiple sub-centers for each class, as shown in Figure \ref{fig2}:
\begin{equation} 
{{w}_{i}}=[{{w}_{i,1}},{{w}_{i,2}},\cdots ,{{w}_{i,s}}],{{w}_{i,j}}\sim \mathsf{N}({{\mu }_{i}},{{\sigma }^{2}})
\label{5} 
\end{equation}
where ${{w}_{i,j}}\in {{R}^{d}}$ denotes the $j$-th sub-center of class $i$. ${{\sigma }^{2}}$ is variance, which is a hyperparameter. To reduce the increase in calculation caused by setting multiple sub-centers, after completing the initialization, we fix ${{w}_{i,j}}$. The parameters of ${{w}_{i,j}}$ do not change during the training process and do not calculate the gradient: ${{\overset{\wedge }{\mathop{w}}\,}_{i,j}}=\mathsf{FIX}({{w}_{i,j}})$, where $\mathsf{FIX}()$ represents stop gradient. Use ${{\overset{\wedge }{\mathop{w}}\,}_{i,j}}$ instead of $w$ in Eq.(\ref{2}):
\begin{equation} 
{{\overset{\sim }{\mathop{p}}\,}_{{{x}_{i,{{y}_{i}},k}}=}}\frac{\exp (\overset{\wedge }{\mathop{w}}\,_{{{y}_{i}},k}^{\mathsf{T}}{{x}_{i}})}{\sum\nolimits_{j=1}^{c}{\sum\nolimits_{m=1}^{s}{\exp (\overset{\wedge }{\mathop{w}}\,_{j,m}^{T}{{x}_{i}})}}}
\label{6} 
\end{equation}

This method brings two main beneﬁts:
\begin{itemize}
\item By setting different ${{\sigma }^{2}}$ and fixing ${{w}_{i,j}}$, F-SC can control the distance between different sub-centers of the same class and the sub-centers will not be close to each other during training, which ensures the ability of different sub-centers to capture different modes of the same class.
\item As we keep the classifier fixed, less parameters need to be updated, reducing the computational cost of the model during training.
\end{itemize}

\subsection{Intra-SubClass Push Force}

Dividing a class into multiple sub-classes may undermine the intra-class compactness. To enhance intra-class compactness, we use a strategy similar to the center loss \cite{wen2016discriminative} to constrain the features of the samples. But, unlike the center loss, which calculates the class's feature center, we directly use $\overset{\wedge }{\mathop{w}}\,$ sampled in Eq.(\ref{5}) as the sub-class's feature center and penalize the distance between the features and their corresponding sub-class centers to achieve intra-subclass compactness (as shown in Figure \ref{fig1} (c), Intra-subclass push force). Hence, a sub-class compactness loss (${{L}_{sc-c}}$) is proposed as below:

\begin{equation} 
{{L}_{sc-c}}=\frac{1}{2}\sum\limits_{i=1}^{n}{\left\| {{\mathbf{x}}_{{{y}_{i}},k}}-{{\mathbf{\overset{\wedge }{\mathop{w}}\,}}_{{{y}_{i}},k}} \right\|}_{2}^{2}\label{7} 
\end{equation}

where ${{\mathbf{\overset{\wedge }{\mathop{w}}\,}}_{{{y}_{i}},k}}$ denotes the fixed class center of the $k$-th sub-class of class ${{y}_{i}}$.

\subsection{Overall  Loss}

Combining Eq.(\ref{3}), Eq.(\ref{6}), and Eq.(\ref{7}), the total loss (F-SC loss, ${{L}_{F-SC}}$) is given as:
\begin{equation} 
{{L}_{F-SC}}=-\frac{1}{n}\sum\limits_{i=1}^{n}{\log (\sum\limits_{k=1}^{s}{{{\overset{\sim }{\mathop{p}}\,}_{{{x}_{i,{{y}_{i}},k}}}}})}+\beta {{L}_{sc-c}}
\label{8} 
\end{equation}
where $\beta $ is a hyperparameter that controls the strength of ${{L}_{sc-c}}$.

\section{Experiments}

In this section, we first describe our experimental dataset and experimental settings, then illustrate the settings of F-SC hyperparameters. Next,  ablation experiments are performed to demonstrate effects of the individual components of F-SC. Finally, we compare our method with some of the current most popular multi-subcenter methods.

\subsection{Experimental Settings}

\noindent \textbf{Dataset}. We evaluate the performance of our method using the top-1 and top-5 accuracies of Caltech-256 \cite{griffin2007caltech} and Mini-ImageNet \cite{krizhevsky2012imagenet}. Caltech-256 has 256 classes with more than 80 images in each class. Mini-ImageNet contains a total of 60,000 images from 100 classes. To validate the proposed F-SC on the fine-grained image classification (FGVC) task, we test it on CUB2011 \cite{wah2011caltech} and Cars-196 \cite{krause2015fine}. CUB-2011 is a bird dataset that consists of 200 bird species and 11788 images. Cars-196 consists of 196 car classes and 16185 images.

\noindent \textbf{Implementation details}. We use the PyTorch toolbox \cite{paszke2019pytorch} to implement our experiments. During training, we used the standard SGD optimizer with a momentum of 0.9 to train all the models. The weight decay is set to 4E-5. The cosine learning schedule \cite{he2019bag} with an initial learning rate of 0.01 is adopted. We use ResNet-50 as the backbone. All experiments are conducted on an NVIDIA GeForce GTX 3080 (10 GB) GPU. The images are cropped to 224 × 224 as the input. Random horizontal flipping and random cropping are used for data augmentation. In fine-grained recognition, we follow the experimental setup of \cite{qian2019softtriple}.

\begin{figure}[h]
  \centering
  \subfloat[]{
     \includegraphics[width=0.31\linewidth]{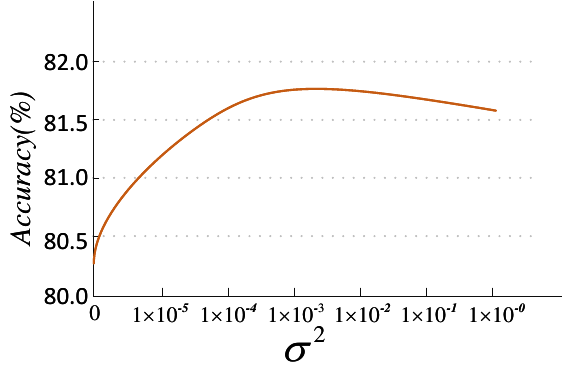}
     \label{set-hyper(a)}
  }
    \subfloat[]{
     \includegraphics[width=0.31\linewidth]{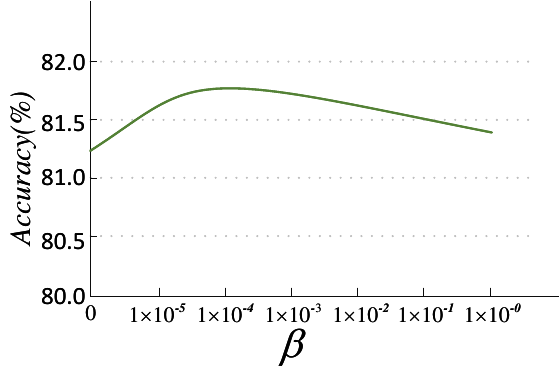}
     \label{set-hyper(b)}
  }
  \subfloat[]{
     \includegraphics[width=0.31\linewidth]{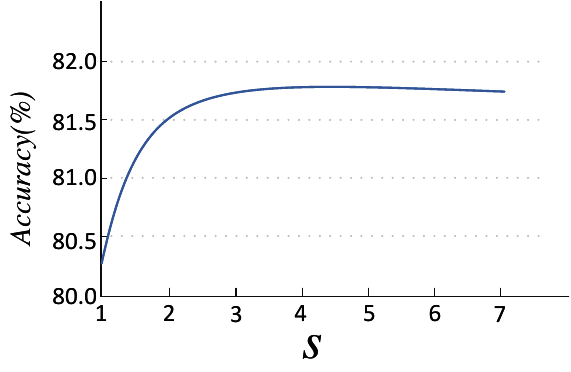}
     \label{set-hyper(c)}
  }\\
  \caption{Performance curve testing on Mini-ImageNet dataset, achieve by (a) models with diﬀerent  ${{\sigma }^{2}}$ and ﬁxed  $\beta =1\times {{10}^{-4}}$,$s=4$.  (b)  models with diﬀerent $\beta $ and ﬁxed ${{\sigma }^{2}}=1\times {{10}^{-3}}$, $s=4$. (c) Accuracies on Mini-ImageNet dataset, models with diﬀerent $s$ and ﬁxed ${{\sigma }^{2}}=1\times {{10}^{-3}}$, $\beta =1\times {{10}^{-4}}$.}
  \label{set-hyper}
\end{figure} 

\subsection{Hyperparameter Settings}

The hyperparameter $s$ sets the number of sub-classes for each class, the hyperparameter ${{\sigma }^{2}}$ controls the distance between subclasses, and $\beta $ dominates the intra-subclass variations. All three parameters are essential to our model. So we conduct three experiments to investigate the sensitiveness of the three parameters. It can be concluded from Figure \ref{set-hyper} (a) that too large ${{\sigma }^{2}}$ may lead to too large a distance between sub-centers, which affects the intra-class compactness and reduces the performance of the model. From Figure \ref{set-hyper} (b), it can be derived that ${{L}_{sc-c}}$ dominates the overall loss function as $\beta $ increases, which also affects the performance of the model. From Figure \ref{set-hyper} (c), it can be concluded that after the number of sub-centers reaches a certain value, the model performance tends to be saturated. In the following experiments we set ${{\sigma }^{2}}=1\times {{10}^{-3}}$, $\beta =1\times {{10}^{-4}}$ and $s=4$.

\subsection{Ablation Studies}

\begin{table}[h]
   \centering
   \caption{Top-1 accuracy (\%) with different F-SC components on Mini-ImageNet and Caltech-256. All models were trained in the same experimental settings.}
   \begin{tabular}{l|c|c}
      \hline 
      Method                   & Mini-ImageNet     & Caltech-256 \\
      \hline 
      \hline 
      ResNet50 (baseline)      & 80.13          & 66.21 \\
      \hline 
      only multi-subcenter             & 80.76          & 66.63 \\
      multi-subcenter + ${{L}_{sc-c}}$  & 80.95 & 66.95 \\
      multi-subcenter + F-SC initialization  & 81.35 & 67.08 \\
     \textbf{F-SC}          & \textbf{81.73}          & \textbf{67.76} \\
      \hline 
   \end{tabular} \\
   \label{Ablation-table}
\end{table}

To validate the effectiveness of our multi-subcenter initialization method and ${{L}_{sc-c}}$, we conduct experiments on Mini-ImageNet and Caltech-256, and the results are shown in Table \ref{Ablation-table}. From Table \ref{Ablation-table}, we can find that without any one component on the F-SC, the performance of the model will degrade, which demonstrates the effectiveness of our multi-subcenter initialization method and ${{L}_{sc-c}}$. Specifically, our multi-subcenter initialization method improves accuracy by 0.59\% and 0.45\%, respectively, while ${{L}_{sc-c}}$ improves by 0.19\% and 0.32\%, respectively. This improvement can be attributed to the fact that the normal distribution assumption for each class is realistic and our multi-subcenter initialization method ensures that the sub-centers do not approach each other during training, while ${{L}_{sc-c}}$ increases the intra-class compactness.

\subsection{CUB-2011 and Cars-196}

To validate the proposed F-SC on the fine-grained image classification (FGVC) task, we test it on two datasets, CUB-2011 and Cars-196. We follow the experimental settings in \cite{qian2019softtriple} to evaluate the performance of features learned by different methods, the top-1 accuracy for the recognition task.

\begin{table}[h]
  \caption{Comparisons for different multi-subcenter methods and our F-SC,
  with ResNet-50 baseline.
  Includes the top-1 accuracies (\%) on CUB-2011 and Cars-196 datasets,
  the increments in parameters (+Params), computation (+FLOPs), and memory usage (+MUs)
  compared to the baseline model.}
  \label{FGVC-table}
  \centering
  \begin{tabular}{l|cc|c|c|c}
     \hline
     Model                 & CUB-2011        & Cars-196        & +Params        & +FLOPs        & +MUs \\
     \hline
     \hline 
     ResNet50 (baseline)    & 81.18          & 89.96            & (97.74M)       & (4.12G)      & (109.68M) \\
      \hline
     Center loss \cite{wen2016discriminative}       & 81.23          & 89.85            & 7.52M          & 0.052G        & 13.75M \\
     Subclass distill \cite{muller2020subclass}                & 81.58          & 91.33            & 14.27M          & 0.097G        &20.73M \\
     NWDP \cite{10.1145/3474085.3475536}        & 81.37          &  91.26            & 15.26M               & 0.081G        & 21.03M \\
     SoftTriple  \cite{qian2019softtriple}                      & 81.52          & 91.58            & 15.31M                & 0.095G        &  21.07M \\
     MSCLDL  \cite{9684999}                     & 81.49          & 91.73            & 15.32M                  & 0.092G        & 21.06M \\
     \textbf{F-SC}                & \textbf{82.45} & \textbf{92.28}   & 15.31M    & 0.006G    & 12.13M \\
     \hline
  \end{tabular}
\end{table}

As the results are shown in Table \ref{FGVC-table}. F-SC achieves the best results on these two FGVC datasets. Compared with SoftTriple, our method still gains 0.93\% and 0.70\%, respectively. Besides achieving a considerable performance improvement, F-SC significantly reduces memory usage and computation during model training compared to other multi-subcenter methods.

\subsection{Mini-ImageNet and Caltech-256}

\begin{table}[h]
   \centering
   \caption{Top-1 and Top-5 accuracies (\%) for ResNet-50 model with different multi-subcenter methods, on Mini-ImageNet datasets. The F-SC follows the best settings in the ablation studies, the other multi-subcenter methods follow the settings in their papers.}
   \begin{tabular}{l|c|c}
      \hline 
      Method                   & Top-1 acc.     & Top-5 acc. \\
      \hline 
      \hline 
      ResNet50 (Baseline)      & 80.13          & 93.49 \\
      \hline 
       Center loss \cite{wen2016discriminative}             & 79.95          & 93.81 \\
      Subclass distill \cite{muller2020subclass} &	80.75  &	93.78\\
      NWDP \cite{10.1145/3474085.3475536} & 81.03  & 93.97 \\
      SoftTriple  \cite{qian2019softtriple} &	80.66 &	93.52\\
      MSCLDL  \cite{9684999}  &	80.45 &	94.55\\
      \textbf{F-SC}& \textbf{81.73}  & \textbf{94.83}\\
      \hline 
   \end{tabular} \\
   \label{Mini-ImageNet (Top-1 and Top-5)}
\end{table}

In this section, we evaluate the performance of our method using top-1 and top-5 accuracy on Mini-ImageNet and Caltech -256. For both datasets, we follow the standard testing protocols. The classification performance of different losses under the same training conditions is listed in Table \ref{Mini-ImageNet (Top-1 and Top-5)} and Table \ref{Caltech-256 (Top-1 and Top-5)}.

The results show that F-SC outperforms other losses under the same experimental settings. Specifically, on the Mini-ImageNet, F-SC achieves a noticeable performance gain of 1.07\% compared to SoftTriple and has more improvements compared to other SoftMax variants. On the Caltech-256 dataset, F-SC also achieves better performance than other methods. Additionally, using only center loss has essentially no effect on model performance. Compared to SoftTriple, the performance of F-SC on top-1 is improved by 1.07\% and 0.58\%, respectively. This shows that F-SC can depict the inherent structure of data better.

\begin{table}[h]
   \centering
   \caption{Results on Caltech-256 (Top-1 and Top-5 acc, \%).}
   \begin{tabular}{l|c|c}
      \hline 
      Method                   & Top-1 acc.     & Top-5 acc. \\
      \hline 
      \hline 
      ResNet50 (Baseline)      & 66.21          & 81.39 \\
      \hline 
       Center loss \cite{wen2016discriminative}             & 66.17          & 81.52 \\
      Subclass distill \cite{muller2020subclass} &	67.03  &	82.36\\
      NWDP \cite{10.1145/3474085.3475536} & 67.25  & 83.49 \\
      SoftTriple  \cite{qian2019softtriple} &	67.18 &	83.93\\
      MSCLDL  \cite{9684999}  &	67.44 &	84.06\\
      \textbf{F-SC}& \textbf{67.76}  & \textbf{84.73}\\
      \hline 
   \end{tabular} \\
   \label{Caltech-256 (Top-1 and Top-5)}
\end{table}

\section{Related Work}

The concept of sub-center has been introduced into image classification for a long time. In \cite{7862266}, to reduce the overlap of sub-classes, this paper proposes an extension of LDA, separability oriented subclass discriminant analysis (SSDA), which employs hierarchical clustering to divide a class into subclasses using a separability oriented criterion.  In \cite{GUO2021104033}, to get rid of the assumption that all samples from the same class are independently identically distributed (i.i.d.), this paper, reverse nearest neighbor (RNN) technique is imbedded into L2BLDA and a novel linear discriminant analysis named RNNL2BLDA is proposed. Rather than using classes to construct within-class and between-class scatters, RNNL2BLDA divides each class into subclasses by using RNN technique. The experimental results in \cite{7862266,GUO2021104033} show that more than one sub-center can effectively capture different face modalities; for example, one sub-center captures the front view, and another sub-center captures the side view.

Softtriple \cite{qian2019softtriple} combines sub-center and SoftMax in fine-grained image classification, and this combination has better performance than using SoftMax alone. These sub-centers help the neural network better capture the different modalities of the data. In \cite{deng2020sub}, the sub-center is used to capture noise samples from face datasets for denoising to improve the performance of face datasets with much noise. NWDP \cite{10.1145/3474085.3475536} proposes to use multi-subcenters to distinguish in- and out-of-distribution noisy samples, and purify the web training data by discarding out-of-distribution noisy images and relabeling in-distribution images for better robustness and performance. Sub-class distill \cite{muller2020subclass,9411995} divides a class into multiple sub-classes and then improves the efficiency of knowledge distillation through these sub-classes of different modalities. MSCLDL \cite{9684999} proposes a novel framework, namely Multi-Subclass Classiﬁcation with Label Distribution Learning , which decomposes two classes into a certain number of sub-classes according to image structure. Based on the multi-subclass training set, a label distribution model is learned to generate more accurate saliency maps in some challenging scenes. Sub-class has been widely used in face recognition \cite{7862266,GUO2021104033}, image classification \cite{qian2019softtriple}, image retrieval \cite{DBLP:journals/corr/abs-2112-11689} and image denoising \cite{9534290, deng2020sub}. However, these methods do not consider the problem of highly overlapping sub-centers, and the extra calculation that comes with setting up multiple sub-centers.

\section{Conclusion}
\label{Conclusion-label}

The use of more than one sub-center can capture different modalities of data. However, sub-centers created by existing methods may be highly overlapping , the classifier's parameters grow linearly with the number of classes and lose the intra-class compactness. The F-SC, a novel perspective of sub-center, is proposed to address these three problems uniformly. This method ensures the diversity of sub-centers and dramatically reduces the amount of calculation during training. Furthermore, a constraint of Euclidean distance is added to the feature of each sub-class, which is used to improve the intra-subclass compactness of the features. F-SC can be easily set up in existing neural networks without complicated network structure modifications. F-SC reduces the computation and memory usage during training, while the amount of computation in the model inference phase is not reduced. How to further reduce the additional computation due to setting multiple sub-centers is our next step. Experiments on Mini-Imagenet, Caltech-256, CUB-2011, and Cars-196 have validated the effectiveness of our proposed F-SC.


{
\small
\bibliographystyle{plainnat}
\bibliography{mulit-center}
}

\clearpage


\end{document}